# The Learning of Fuzzy Cognitive Maps With Noisy Data: A Rapid and Robust Learning Method With Maximum Entropy

Guoliang Feng, Wei Lu, Witold Pedrycz, *Fellow, IEEE*, Jianhua Yang, and Xiaodong Liu

*Abstract*—Numerous learning methods for fuzzy cognitive maps (FCMs), such as the Hebbian-based and the population-based learning methods, have been developed for modeling and simulating dynamic systems. However, these methods are faced with several obvious limitations. Most of these models are extremely time consuming when learning the large-scale FCMs with hundreds of nodes. Furthermore, the FCMs learned by those algorithms lack robustness when the experimental data contain noise. In addition, reasonable distribution of the weights is rarely considered in these algorithms, which could result in the reduction of the performance of the resulting FCM. In this article, a straightforward, rapid, and robust learning method is proposed to learn FCMs from noisy data, especially, to learn large-scale FCMs. The crux of the proposed algorithm is to equivalently transform the learning problem of FCMs to a classic-constrained convex optimization problem in which the least-squares term ensures the robustness of the well-learned FCM and the maximum entropy term regularizes the distribution of the weights of the well-learned FCM. A series of experiments covering two frequently used activation functions (the sigmoid and hyperbolic tangent functions) are performed on both synthetic datasets with noise and real-world datasets. The experimental results show that the proposed method is rapid and robust against data containing noise and that the well-learned weights have better distribution. In addition, the FCMs learned by the proposed method also exhibit superior performance in comparison with the existing methods.

Manuscript received April 15, 2019; revised June 17, 2019; accepted August 1, 2019. This work was supported in part by the Natural Science Foundation of China under Grant 61876029 and Grant 61673082, in part by the Canada Research Chair, and in part by the Natural Sciences and Engineering Research Council of Canada. This article was recommended by Associate Editor S.-M. Chen. *(Corresponding author: Wei Lu.)*

G. Feng is with the School of Control Science and Engineering, Dalian University of Technology, Dalian 116023, China, and also with the School of Automation Engineering, Northeast Electric Power University, Jilin 132012, China (e-mail: fengguoliang911@foxmail.com).

W. Lu, J. Yang, and X. Liu are with the School of Control Science and Engineering, Dalian University of Technology, Dalian 116023, China (e-mail: luwei@dlut.edu.cn).

W. Pedrycz is with the Department of Electrical and Computer Engineering, University of Alberta, Edmonton, AB T6R 2V4, Canada, also with the Systems Research Institute, Polish Academy of Sciences, 00-901 Warsaw, Poland, and also with the Department of Electrical and Computer Engineering, Faculty of Engineering, King Abdulaziz University, Jeddah 21589, Saudi Arabia (e-mail: wpedrycz@ualberta.ca).

## I. Introduction

THE FUZZY cognitive maps (FCMs), which were introduced by Kosko [1], are widely applied to knowledge-based representation and reasoning, complex system modeling [2]–[4]; time-series forecasting [5], [6]; decision making [7]–[9]; classification patterns [10]–[12]; risk assessment and management [13], [14]; and others. The FCMs are the directed graphs with feedback loops, which consists of nodes and directed weights. Here, the nodes represent some concepts, variables, or key factors that are encountered in the system while the directed weights reflect the degrees of interaction between those nodes [15].

When the structure of an FCM is given in advance, how to learn the weights between the individual nodes according to the experimental data is an ongoing timely issue. Over the years, researchers have presented diverse approaches to address this issue. In general, the learning methods of FCMs fall into the three categories [16], that is, Hebbian-based, population-based, and hybrid learning methods.

Hebbian-based learning methods are a kind of simple and fast method for learning FCMs. Their fundamental idea is to iteratively adjust weights of FCMs based on the Hebbian formula that is frequently used in the learning artificial neural networks and available data until some stopping criteria are reached. This category of learning methods of FCMs mainly includes differential Hebbian learning (DHL) [17] algorithm, nonlinear Hebbian learning (NHL) [18], active Hebbian learning (AHL) [19], data-driven NHL (DDNHL) [20], etc.

Population-based methods form a category of mainstream FCM learning methods that are widely used at present. They learn weights of FCMs by exploiting some existing population-based optimization technologies to minimize a certain objective function that is expressed in the sum of squared errors of individual nodes *vis-à-vis* experimental data. Here, some population-based optimization technologies, such as the genetic algorithm (GA) [21], particle swarm optimization (PSO) [22], real-coded GA (RCGA) [23], differential evolution (DE) [24], big bang-big crunch (BB-BC) [25], artificial bee colony (ABC) [26], and imperialist competitive algorithm (ICA) [27] are commonly considered. Recently, some improved population-based methods have also been developed to deal with the learning problem of FCMs. Stach *et al.* [28] proposed a sparse RCGA (SRCGA) to learn the weights of FCMs with sparse structures in which a density parameter is added into the objective function for guiding the learning toward the formation of maps with a certain predefined density. Wu and Liu [29] first introduced Lasso regularization as the sparsity penalty term into the objective function to ensure the sparse structure of the resulting FCM and, then, the weights of the FCM were determined by using the evolutionary algorithms to minimize the objective function. To use FCMs to reconstruct gene regulatory networks (GRNs), Zou and Liu [30] proposed the memetic algorithm with mutual information (MIMA-FCM) to optimize the objective function to realize the learning of large-scaled FCMs where mutual information is used to reduce the search space of weights. To handle the causal network reverse engineering problem, Chen et al. [31] exploited the decomposed framework and RCGA to construct the FCMs to simulate GRNs with 300 nodes and 439 edges using gene expression data. To reconstruct the GRNs with high accuracy, a dynamic multiagent GA (dMAGA) is also proposed to simulate a GRN with 200 nodes [32].

The last category of methods, the hybrid methods, combines both the Hebbian-based and population-based learning methods to learn weights of FCMs. Papageorgiou and Groumpos [33] combined the DE algorithm with the NHL algorithm to learn FCMs. In this method, the NHL algorithm is used to build the initial architecture of the FCM using experimental data and, then, the DE algorithm is used to further refine the weights. Zhu and Zhang [34] proposed a hybrid framework by integrating the RCGA and NHL algorithm, and successfully implemented it in the real-world partner selection problems. Liu and Zhang [35] introduced the least-squares method and the Hebbian learning technique to solve the truck backer-upper control problem. Natarajan et al. [36] combined the key aspects of DDNHL and GA to learn the weights of FCMs to accomplish sugarcane yield classification.

The aforementioned methods have contributed to the learning problem of the FCMs. However, they still encounter several evident limitations. One is that those learning algorithms lack robustness since the experimental data contain noise. Another is that many of those algorithms, especially, the most used population-based methods, are extremely time consuming with high-computational overloads when learning the FCMs with hundreds of nodes. The overall number of weights of the FCM to be learned quadratically increases with the number of nodes, which implies that it becomes fairly difficult to find the weights of an FCM with a vast number of nodes. The last one is that the distribution of the weights of the FCM is rarely considered in those learning methods. The weights that are learned by the above-mentioned learning methods are unreasonably distributed in between $-1$ and 1, which could result in the reduction of the performance of the resulting FCM.

To reduce the aforementioned limitations that are encountered in the existing learning methods of FCMs, in this article, a straightforward, rapid, and robust method is proposed to learn the FCMs from noisy data, and it is referred to as *LEFCM*. In the proposed method, the learning problem of the FCMs is equivalently converted into a classic convex optimization problem with constraints. In comparison with the existing learning methods of the FCMs, the advantages of the proposed one are as follows.

1) The LEFCM can learn the weights of FCMs from the large-scale experimental data containing the noise efficiently and rapidly since the proposed method converts the FCM learning problem to a constrained convex optimization problem that can be solved in polynomial time complexity by applying the gradient-based method.
2) The LEFCM exhibits significant robustness in the presence of noisy data since the least-squares term is introduced into the convex optimization problem.
3) The FCM that is generated by the LEFCM has a more reasonable weight distribution since the maximum entropy constraint is also added into the convex optimization problem.

The remainder of this article is organized as follows. Section II presents an outline of the learning problem of FCMs. In Section III, the details of the LEFCM are thoroughly presented. In Section IV, we exploit the synthetic and real data with noise to validate the feasibility and effectiveness of the proposed method. Some comparative results with other existing methods are also covered in this section. Finally, Section V concludes this article and highlights some future studies.

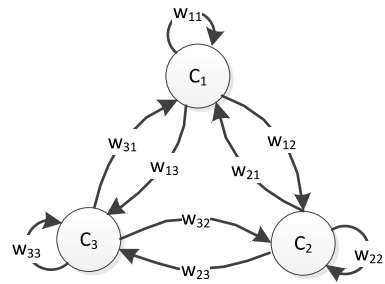

Fig. 1. Example of an FCM consisting of three nodes.

## II. FCM AND ITS LEARNING

The FCMs are rooted from the cognitive maps [37], which are generated by admitting the fuzzy degrees of causality between the individual concepts in the cognitive maps. In brief, an FCM is a weighted digraph with feedback loops, which can depict the dynamic behavior of a system to be investigated. Fig. 1 displays an example of an FCM including three concept nodes.

Generally, an FCM consists of concept nodes and weighted edges. The concept nodes, which represents the variables in the system to be investigated, are denoted as $C_1, C_2, \ldots, C_n$, where $n$ is the total number of concept nodes. The concept nodes in the FCM are connected with directed and weighted edges. The edges reflect causality presented between nodes, or more precisely, a way in which one node affects another one. The edge weight from nodes $c_j$ to $c_i$ is denoted by $w_{ji}, i, j = 1, 2, \ldots, n$, and quantified by numeric values positioned in the interval $[-1, 1]$. The sign and absolute value of the edge weight reflect the influential direction and affected degree of the causality between the nodes, respectively. More specifically:

1) $w_{ji} > 0$ means that an increase of node $C_j$ leads to an increase of node $C_i$ with strength $w_{ji}$ and vice-versa;
2) $w_{ji} < 0$ means that an increase of node $C_j$ leads to a decrease of node $C_i$ with strength $|w_{ji}|$ and vice-versa;
3) $w_{ji} = 0$ means that there is no relationship between nodes $C_j$ and $C_i$.

All edge weights in an FCM can also be stored in a matrix format, that is,

$$\mathbf{W} = \begin{bmatrix} w_{11} & w_{12} & \cdots & w_{1n} \\ w_{21} & w_{22} & \cdots & w_{2n} \\ \vdots & \vdots & \ddots & \vdots \\ w_{n1} & w_{n2} & \cdots & w_{nn} \end{bmatrix} = [\mathbf{w}_1, \mathbf{w}_2, \ldots, \mathbf{w}_n]$$

where $\mathbf{W}$ is called as the weight matrix of the FCM.

The reason that FCM can describe the dynamic behavior of the system to be investigated is that the value of its individual concept nodes vary with time. The state value of each node at $t+1$ time moment is governed by the following formula:

$$A_i(t+1) = f\left(\sum_{j=1}^{n} A_j(t) w_{ji}\right), i = 1, 2, \ldots, n \quad (1)$$

where $A_j(t)$ is the state value of the $j$th node $C_j$ at time $t$; $A_i(t+1)$ is the state value of the $i$th node $C_i$ at time $t+1$; $w_{ji}$ is the edge weight from the node $C_j$ to the node $C_i$; and $f(\cdot)$, called as the activation function, is a monotonically nonlinear continuous nondecreasing function that converts the weighted sum of the state values of all nodes in the FCM into a certain interval. There are two forms of the activation functions that are frequently used in most applications. One is the unipolar sigmoid function that is expressed in the form

$$f(x) = \frac{1}{1 + e^{-\lambda x}} \quad (2)$$

which is used to quantify the state value of the individual nodes into the interval [0, 1]. Another is the hyperbolic tangent function expressed as

$$f(x) = \tanh(\lambda x) = \frac{e^{\lambda x} - e^{-\lambda x}}{e^{\lambda x} + e^{-\lambda x}} \quad (3)$$

which is used to quantify the state value of the individual nodes into the interval [−1, 1]. Note that the parameter $\lambda > 0$ in (2) and (3) controls the shape of the corresponding activation function.

Let the state values of all nodes at time $t$ be a state vector $\mathbf{A}(t) = [A_1(t), A_2(t), \ldots, A_n(t)]$. Thus, (1) is rewritten in a vector format

$$A_i(t+1) = f(\mathbf{A}(t) \mathbf{w}_i) \quad (4)$$

where $\mathbf{w}_i = [w_{1i}, w_{2i}, \ldots, w_{ni}]^T$ is the $i$th column of the weight matrix $\mathbf{W}$. Once the weight matrix $\mathbf{W}$ of an FCM has been given, the FCM starts from a prespecified initial state vector to carry out the consecutive iterations' computation in light of (1) or (4), this may give rise to a variety of dynamic behavior of the FCM, such as oscillations (say, a limit cycle) or convergence to some equilibrium.

The learning problem of an FCM is to determine its weight matrix $\mathbf{W}$ so that the target sequences can be accurately reproduced by the FCM coming with this weight matrix $\mathbf{W}$. In general, the minimization of the following objective function is considered to deal with this problem:

$$\arg\min_{\mathbf{W}}: J = \frac{1}{mnk} \sum_{s=1}^{m} \sum_{t=1}^{k} \sum_{i=1}^{n} \left(A_i^s(t) - \hat{A}_i^s(t)\right)^2$$
$$\text{s.t. } |w_{ji}| \leq 1, \ i, j = 1, 2, \ldots, n \quad (5)$$

where $m$ is the overall number of different initial state vectors, $n$ is the number of nodes in the FCM to be learned, and $k$ is the length of the response sequences. Moreover, in (5), $A_i^s(t)$ represents the target state value of the $i$th node $C_i$ at time $t$ for the $s$th initial state vector, and $\hat{A}_i^s(t)$ expresses the estimated value of the $i$th node $C_i$ produced by the FCM with the weight matrix $\mathbf{W}$ at time $t$ for the $s$th initial state vector, where $s = 1, 2, \ldots, m$. Currently, many learning methods were developed to minimize (5). However, in what follows, a novel method is presented to transform the learning problem of the FCM to the classic convex optimization problem with constraints instead of the nonconvex problem that is expressed as (5).

## III. Proposed Learning Method of FCM

In this section, a novel learning method of the FCM is presented in detail. Its crux lies in equivalently transforming the learning problem of the FCM into a classic convex optimization problem with constraints. Suppose that there is a dynamical system including $n$ concepts (variables). To investigate the system, $m$ different initial state vectors $[x_{01}^s, x_{02}^s, \ldots, x_{0n}^s]$ with $s = 1, 2, \ldots, m$ are fed into this system to stimulate the states of these concepts within a limited time $k$. For each initial state vector, this system produces $n$ response sequences including $k$ state values. Organizing these response sequences in a matrix format, we finally form a total of $m$ state response matrices

$$\mathbf{D}_s = \begin{bmatrix} x_{11}^s & x_{12}^s & \cdots & x_{1n}^s \\ x_{21}^s & x_{22}^s & \cdots & x_{2n}^s \\ \vdots & \vdots & \vdots & \vdots \\ x_{k1}^s & x_{k2}^s & \cdots & x_{kn}^s \end{bmatrix}, s = 1, 2, \ldots, m$$

where $m$ represents the number of different initial state vectors, $n$ is the number of concepts in the system to be investigated, and $k$ is the length of the response sequences. The $i$th column of each state response matrix is a response sequence in regard to the $i$th concept in the system under different initial state vectors. In other words, these state response matrices depict the dynamic behavior of the system to be investigated in the context of different initial conditions. Now, we use an FCM consisting of $n$ nodes to map those dynamic behaviors. Here, the nodes of the FCM agree with the concepts of the system to be investigated. Our task is to determine a suitable weight matrix $\mathbf{W}$ so that the target sequences $\mathbf{D}_1, \mathbf{D}_2, \ldots, \mathbf{D}_m$ can be reproduced by the FCM with this weight matrix under the stimulation of those $m$ initial state vectors.

In order to complete the above task, we first consider the $i$th node in the FCM and its target sequence $x_{1i}^s, x_{2i}^s, \ldots, x_{ki}^s$ with respect to the $s$th initial state vector $[x_{01}^s, x_{02}^s, \ldots, x_{0n}^s]$. When the state values of individual concepts (nodes) of the system observed at $t$ time moment under the $s$th initial state vector, which is denoted by a vector $\mathbf{X}_s(t) = [x_{t1}^s, x_{t2}^s, \ldots, x_{tn}^s]$, are fed into the FCM, the state value of the $i$th node at $t+1$ time moment can be obtained by (4), viz.,

$$A_i(t+1) = f(\mathbf{X}_s(t)\mathbf{w}_i). \qquad (6)$$

For each $t = 0, 1, \ldots, k-1$, we envision that $A_i(1), A_i(2), \cdots, A_i(k)$ become $x_{1i}^s, x_{2i}^s, \ldots, x_{ki}^s$ in order, which means that there is no difference between the response sequences of the $i$th node produced by feeding the $s$th initial state vector to this FCM and the corresponding target sequences coming with this initial state vector. Thus, we have

$$x_{(t+1)i}^s = f(\mathbf{X}_s(t)\mathbf{w}_i) \qquad (7)$$

with $t = 0, 1, \ldots, k-1$. By noting that the activation function $f(\cdot)$ is a monotonically nonlinear continuous nondecreasing function, its inverse function $f^{-1}(\cdot)$ can be uniquely determined. Here, for the unipolar sigmoid, its inverse function $f_{\text{sigmoid}}^{-1}$ is

$$f_{\text{sigmoid}}^{-1}(y) = -\frac{1}{\lambda}\ln\frac{1-y}{y} \qquad (8)$$

whereas for the hyperbolic tangent function, its inverse function $f_{\text{tanh}}^{-1}$ is

$$f_{\text{tanh}}^{-1}(y) = \frac{1}{2\lambda}\ln\frac{1+y}{1-y}. \qquad (9)$$

Thus, (7) can be rewritten as

$$f^{-1}\left(x_{(t+1)i}^s\right) = \mathbf{X}_s(t)\mathbf{w}_i \qquad (10)$$

with $t = 0, 1, \ldots, k-1$.

In light of (10), for all $s = 1, 2, \ldots, m$, a system of linear equations including $mk$ equations and $n$ unknown variables $w_{1i}, w_{2i}, \ldots, w_{ni}$ is produced, say

$$\mathbf{D}_1 \begin{cases} \begin{bmatrix} f^{-1}(x_{1i}^1) \\ f^{-1}(x_{2i}^1) \\ \vdots \\ f^{-1}(x_{ki}^1) \end{bmatrix} \\ \end{cases}$$
$$\mathbf{D}_2 \begin{cases} \begin{bmatrix} f^{-1}(x_{1i}^2) \\ f^{-1}(x_{2i}^2) \\ \vdots \\ f^{-1}(x_{ki}^2) \end{bmatrix} \\ \end{cases} = \begin{bmatrix} x_{01}^1 & x_{02}^1 & \cdots & x_{0n}^1 \\ x_{11}^1 & x_{12}^1 & \cdots & x_{1n}^1 \\ \vdots & \vdots & \ddots & \vdots \\ x_{(k-1)1}^1 & x_{(k-1)2}^1 & \cdots & x_{(k-1)n}^1 \\ x_{01}^2 & x_{02}^2 & \cdots & x_{0n}^2 \\ x_{11}^2 & x_{12}^2 & \cdots & x_{1n}^2 \\ \vdots & \vdots & \ddots & \vdots \\ x_{(k-1)1}^2 & x_{(k-1)2}^2 & \cdots & x_{(k-1)n}^2 \\ \vdots & \vdots & \ddots & \vdots \\ x_{01}^m & x_{02}^m & \cdots & x_{0n}^m \\ x_{11}^m & x_{12}^m & \cdots & x_{1n}^m \\ \vdots & \vdots & \ddots & \vdots \\ x_{(k-1)1}^m & x_{(k-1)2}^m & \cdots & x_{(k-1)n}^m \end{bmatrix} \begin{bmatrix} w_{1i} \\ w_{2i} \\ \vdots \\ w_{ni} \end{bmatrix}$$
$$\mathbf{D}_m \begin{cases} \begin{bmatrix} f^{-1}(x_{1i}^m) \\ f^{-1}(x_{2i}^m) \\ \vdots \\ f^{-1}(x_{ki}^m) \end{bmatrix} \\ \end{cases}$$

where $|w_{ji}| \leq 1, j = 1, 2, \ldots, n$. Further, the above system of linear equations can be simplified as

$$\mathbf{Y}_i = \mathbf{X}\mathbf{w}_i$$
$$\text{s.t.} \quad |w_{ji}| \leq 1, j = 1, 2, \ldots, n \qquad (11)$$

where $\mathbf{Y}_i$ is an $mk$-by-1 matrix that is associated with all response values that are generated by the $i$th node (concept) in the condition of $m$ initial states, and $\mathbf{X}$ is an $mk$-by-$n$ matrix with respect to the response values that are generated by all nodes (concepts) in the condition of $m$ initial states except for the last time step. The vector $\mathbf{w}_i$ is the $i$th column of the weight matrix $\mathbf{W}$ of the FCM to be learned.

Obviously, the solution of the system of linear equations with constraints in (11) can make the value of (5) become zero. However, the system of linear equations with constraint could have no solution due to the experimental data containing noise. Alternatively, $\mathbf{w}_i$ can be well approximated by minimizing the error $\|\mathbf{X}\mathbf{w}_i - \mathbf{Y}_i\|_2$. Thus, (11) is converted to the following least-squares problem with constraints:

$$\arg\min_{\mathbf{w}_i}: \|\mathbf{X}\mathbf{w}_i - \mathbf{Y}_i\|_2$$
$$\text{s.t.} \quad \|\mathbf{w}_i\|_\infty \leq 1 \qquad (12)$$

where $\|\mathbf{w}_i\|_\infty = \max\{|w_{1i}|, |w_{2i}|, \ldots, |w_{ni}|\}$ and $\|\mathbf{X}\mathbf{w}_i - \mathbf{Y}_i\|_2 = [\sum_{s=1}^m \sum_{j=0}^{k-1} [\sum_{l=1}^n x_{jl}^s w_{li} - f^{-1}(x_{(j+1)i}^s)]^2]^{(1/2)}$. Note that in (12), the constraint condition $\|\mathbf{w}_i\|_\infty \leq 1$ guarantees that the individual entries of the solution $\mathbf{w}_i$ are located in the interval $[-1, 1]$.

Further, we take into account that the weight matrix of the large-scale FCM is commonly sparse [28], which requires that the 1-norm of weights, $\|\mathbf{w}_i\|_1$, is as small as possible. In addition, we expect that the well-learned weights exhibit a better distribution, which requires that the entropy of the probability distribution of weights, $H(P(\mathbf{w}_i))$, is as large as possible. To realize these two appeals, we add $\|\mathbf{w}_i\|_1$ and $H(P(\mathbf{w}_i))$ as two penalty items into (12). Thus, the learning problem of FCM is completely converted to the following optimization problem with constraints:

$$\arg\min_{\mathbf{w}_i}: \|\mathbf{X}\mathbf{w}_i - \mathbf{Y}_i\|_2 + \beta\|\mathbf{w}_i\|_1 - \alpha H(P(\mathbf{w}_i))$$
$$\text{s.t.} \quad \|\mathbf{w}_i\|_\infty \leq 1 \qquad (13)$$

where $\|\mathbf{w}_i\|_1 = \sum_{j=1}^n |w_{ji}|$ and $\alpha, \beta > 0$ are regularization parameters. Suppose that the probability distribution $P$ assigns a non-negative probability $p(w_{ji})$ to each entry $w_{ji}$ ($j = 1, 2, \ldots, n$) in the vector $\mathbf{w}_i$. The entropy of the weights of the edges that are connected to the $i$th node $C_i$ is expressed in the form

$$H(P(\mathbf{w}_i)) = -\sum_{j=1}^n p(w_{ji}) \log p(w_{ji}) \qquad (14)$$

where $n$ is the total number of nodes of the FCM to be learned. However, it is difficult to solve this entropy problem because the probability distribution $P$ is unknown. To simplify the numerical calculations, the element $(w_{ji} + 1)/2$ is used to replace the probability $p(w_{ji})$ in (14), say

$$H(P(\mathbf{w}_i)) = \widetilde{H}(P(\mathbf{w}_i)) = -\sum_{j=1}^n \frac{w_{ji}+1}{2} \log \frac{w_{ji}+1}{2}. \qquad (15)$$

**Algorithm 1** LEFCM
---
**Input:**
1: $\mathbf{D}_1, \mathbf{D}_2, \cdots, \mathbf{D}_m$: $m$ target sequences, each of these sequences is a $k$-by-$n$ matrix;
2: *type*: the type of activation function;
3: $\lambda$: the shape parameter of activation function;
4: $\alpha$: the parameter of entropy penalty term;
5: $\beta$: the parameter of $L_1$ regularization;

**Output:**
6: $\mathbf{W}$: the well-learned weight matrix.
7: $\mathbf{W} \leftarrow \mathbf{0}^{n \times n}$
8: $\mathbf{X} \leftarrow [\mathbf{D}_1[1:k-1, :]; \mathbf{D}_2[1:k-1, :]; \cdots ; \mathbf{D}_m[1:k-1, :]]$
9: **If** *type* is sigmoid function **Then**
10: $\quad f^{-1}(\cdot)$ is described as (8)
11: **Else If** *type* is hyperbolic tangent function **Then**
12: $\quad f^{-1}(\cdot)$ is described as (9)
13: **End If**
14: $\mathbf{Y}_i = [f^{-1}(\mathbf{D}_1[2:k, i]); f^{-1}(\mathbf{D}_2[2:k, i]); \cdots ; f^{-1}(\mathbf{D}_m[2:k, i])]$
15: $i \leftarrow 1$
16: **While** $i \leq n$ **Do**
17: $\quad$ Determine $\mathbf{w}_i$ by directly invoking existing convex optimization methods to solve (16)
18: $\quad \mathbf{W}[:, i] \leftarrow \mathbf{w}_i$
19: $\quad i \leftarrow i + 1$
20: **End While**

Thus, (13) can be rewritten as follows:

$$\arg\min_{\mathbf{w}_i}: \quad \|\mathbf{X}\mathbf{w}_i - \mathbf{Y}_i\|_2 + \beta\|\mathbf{w}_i\|_1 - \alpha\widetilde{H}(P(\mathbf{w}_i))$$
$$\text{s.t.} \quad \|\mathbf{w}_i\|_\infty \leq 1. \quad (16)$$

In (16), its first term, which estimates the weights by minimizing (12), ensures that the candidate solution is less sensitive to noise. The second one makes the solution become sparse, that is, the resulting $\mathbf{w}_i$ has more zero entries. With the increase of $\beta$, the number of nonzero entries in $\mathbf{w}_i$ decreases. The last one ensures that the distribution of weights is more reasonable rather than being indiscriminately crowded in its well-defined interval $[-1, 1]$. The reason behind it is that the maximum entropy of a random variable results in generating the best probability distribution of the variable [38].

Since individual items in (16) are all convex, (16) is a classic convex optimization problem with constraints in essence. It can be efficiently solved by many convex optimization techniques, such as the barrier function method, the primal–dual method [39], or their improved versions, etc. Once the iterative solution process of (16) has been completed, a global optimal solution $\mathbf{w}_i^*$ with respect to the $i$th column of the weight matrix $\mathbf{W}$ to be learned can be finally determined. The remaining columns of this weight matrix can also be determined in the same way through repeating the above solving process $n$ times for $n$ nodes. So far, the weight matrix of the well-learned FCM is finally obtained, that is, $\mathbf{W}^* = [\mathbf{w}_1^*, \mathbf{w}_2^*, \ldots, \mathbf{w}_n^*]$. Algorithm 1 depicts the overall procedure of the proposed LEFCM.

TABLE I
PARAMETERS OF PSO USED IN ALL EXPERIMENTS

| Parameters | Values |
|---|---|
| Ranges of input variables | $[-1, 1]$ |
| Population size | 20 |
| Maximum number of iterations | 500 |
| Acceleration const 1 | 2 |
| Acceleration const 2 | 2 |
| Initial inertia weight | 0.9 |
| Final inertia weight | 0.4 |
| Minimum global error gradient | $1e^{-20}$ |

## IV. EXPERIMENTAL STUDIES

In this section, both synthetic data and real-world data are used to test and evaluate the performance of the proposed LEFCM, and to complete the comparison with other existing learning methods. For a fair comparison, all of the tests are implemented on a regular laptop with a 2.3-GHz CPU speed and a 4G of memory using MATLAB R2018a. In this article, the primal–dual interior point method [39], [40] is directly invoked to solve the constrained convex optimization problem in (16).

In all experiments, leave-one-out cross-validation is used to guarantee the credibility of the experimental results. The learning process is repeated $m$ times to calculate the average values and standard deviations of the evaluation metrics for both the synthetic noisy data and real data consisting of $m$ state response matrices. A systematic comparison is made between the proposed LEFCM and the other two representative learning algorithms for FCMs, namely, NHL and PSO. For NHL, the initial weight matrix is randomly generated and the learning rate is set to 0.04. Whereas for PSO, its parameters are listed in Table I. Some other parameters in NHL and PSO are also tested in the experiments but no visible improvement of the fitness values of (5) was reported.

### A. Evaluation Metrics

There are five metrics that are used to evaluate the performance of the proposed learning method. They are defined as follows.

The *Data error* is used to evaluate the ability of the well-learned FCM to reconstruct the experimental data, and it is defined as the mean-squared error between the training sequence and the sequence that is generated by simulating the well-learned FCM from the same initial state vector. The measure of this metric has already presented in (5). The smaller the *Data error*, the stronger the reconstruction ability of the well-learned FCM.

The *Out of sample error* measures the generalization ability of the well-learned FCM [23]. To calculate this metric, both the target FCM and the well-learned FCM are simulated from $m$ randomly chosen initial state vectors that are different from the initial vectors for learning the FCM. Then, we calculate the average absolute values of the differences between the response sequences that are generated by the target FCM and the response sequences that are generated by the well-learned FCM. The smaller the *Out of sample error*, the better the generalization ability of the well-learned FCM

*Out of sample error*
$$= \frac{1}{mnk} \sum_{s=1}^{m} \sum_{t=1}^{k} \sum_{i=1}^{n} |A_i^s(t) - \hat{A}_i^s(t)|. \quad (17)$$

The *Model error* evaluates the similarity between the weights of the target FCM and the ones of the well-learned FCM. The smaller the *Model error* is, the more similar the target FCM and the well-learned FCM are

$$\text{Model error} = \frac{1}{nn} \sum_{j=1}^{n} \sum_{i=1}^{n} |w_{ij} - \hat{w}_{ij}|. \quad (18)$$

The SS Mean, which is the mean value of the specificity and sensitivity [41], is used to evaluate the existence of links between nodes in the well-learned FCM and the target one. To calculate this metric, the weight matrices of the two FCMs are required to be converted into the binary ones where the weights whose absolute values are larger than 0.05 are set to 1 and otherwise are set to 0 [41]. This metric is defined as

$$\text{SS Mean} = \frac{2 \times \text{Specificity} \times \text{Sensitivity}}{\text{Specificity} + \text{Sensitivity}} \quad (19)$$

where Specificity = (TN/TN + FP) and Sensitivity = (TP/TP + FN). TP is the number of zero weights of the target FCM that are correctly identified as zero in the well-learned FCM, and TN is the number of nonzero weights of the target FCM that are correctly identified as nonzero in the well-learned FCM. FP is the number of nonzero weights of the target FCM that are incorrectly identified as zero in the well-learned FCM, and FN is the number of zero weights of the target FCM that are incorrectly identified as nonzero in the well-learned FCM. The value of SS Mean ranges between 0 and 1. The higher the value of it is, the greater the similarity between the well-learned FCM and the target FCM.

The *Execution time*, which is expressed in seconds, evaluates the running time of a given learning method that is used to learn the weight matrix of an FCM. This metric does not include the time to load data and to evaluate model.

It is worth noting that in the above-mentioned metrics, the *Model error* and SS Mean are used for illustrative purposes only since an actual weight matrix could be never known in many real-world problems.

### B. Synthetic Noisy Data

For testing and comparison purposes, the simulated FCMs with varying scales are first constructed and, then, the synthetic noisy data are generated by the FCMs before starting the experiments.

To construct a simulated FCM, some of the weights of the FCM including $n$ nodes are set to nonzero values on the basis of the predefined density, for example, 20% of weights are set to nonzero values corresponding to 20% density, and these nonzero weights are assigned to random values from $[-1, 1]$, of which, the nonzero weight whose absolute value is less than 0.05 is set to 0. In this way, the structure of the simulated FCM is established. Subsequently, $m$ initial state vectors are randomly generated in the intervals [0, 1] (if the activation function is a sigmoid function) or $[-1, 1]$ (if the activation function is a tanh function), and fed into the simulated FCM. Thus, the corresponding synthetic data including $m$ response sequences are generated by implementing the iterative computing from each initial state vector in terms of (1) until $k$ iterative steps are reached. So far, the synthetic data without noise are generated. Further, these generated data are affected by a Gaussian noise $N(\mu, \sigma)$, where $\mu$ represents the mean value and $\sigma$ stands for the standard deviation. In this way, the synthetic data with noise are completely produced. These data are used to test the robustness of the proposed learning algorithm against noise.

TABLE II
SIMULATED FCMS USED IN EXPERIMENTS

| Name | #Nodes | $\lambda_{real}$ sigmoid | $\lambda_{real}$ tanh | Density | $m \times k$ |
|---|---|---|---|---|---|
| C20 | 20 | 5 | 1 | 20% | $5 \times 100$ |
| C40 | 40 | 5 | 1 | 40% | $10 \times 40$ |
| C100 | 100 | 0.7 | 0.8 | 30% | $5 \times 20$ |
| C200 | 200 | 0.2 | 0.4 | 30% | $10 \times 10$ |

Table II lists the simulated FCMs that are used in the experiments, where $\lambda_{real}$ is the value of parameter $\lambda$ of the activation function that is used when generating some simulated FCM, $m$ is the number of initial state vectors, and $k$ is the length of the response sequences that are generated per initial state vector.

There are three hyper-parameters in (16), that is, the shape parameter of activation function $\lambda$, and the regularization parameters $\alpha$ and $\beta$. The values of these parameters have important influences on the learning process of the LEFCM. In this article, the random search (RS) method [42] is exploited to experimentally determine these three parameters. In more detail, for a certain simulated FCM, these three parameters are first randomly generated in their respective search space, which results in the formation of a series of combinations of parameters. Then, the proposed algorithm with those parameter combinations is invoked to learn this simulated FCM. Subsequently, a combination of parameters that minimizes the *Data error* of candidate FCMs is selected as the optimal values of those three parameters, say, $\lambda_{\text{opt}}$, $\alpha_{\text{opt}}$, and $\beta_{\text{opt}}$. In all experiments, the total number of randomly generated combinations of parameters is set to 200. The search spaces range in (0, 0.3) for parameter $\alpha$, (0, 0.5) for parameter $\beta$, and (0, 5.5) for parameter $\lambda$.

Tables III and IV report the values of the shape parameter $\lambda$ that is determined by implementing the RS method on synthetic data without noise and with Gaussian noise $N(0, 0.01)$ and $N(0, 0.1)$ in the case of the sigmoid activation function and the tangent activation function, respectively. For synthetic data without noise, the values of the shape parameter that are sought by the RS method are very close to the $\lambda_{\text{real}}$ that is predefined in the simulated FCMs. For synthetic data with noise, the values of the shape parameter that are sought by the RS method are also closely approximate to the predefined $\lambda_{\text{real}}$ for small-scale FCMs, whereas for large-scale FCMs, the values of the shape parameters that are sought by the RS method exhibit some differences from the predefined $\lambda_{\text{real}}$ for these two activation functions.

TABLE III
SYNTHETIC DATA: THE SHAPE PARAMETER OF THE SIGMOID FUNCTION DETERMINED BY THE RS METHOD

| Data | $\lambda_{real}$ | $\lambda_{opt}$ | | |
|---|---|---|---|---|
| | | Without noise | $N(0, 0.01)$ | $N(0, 0.1)$ |
| C20 | 5 | 5.02 | 5.03 | 5.40 |
| C40 | 5 | 4.98 | 5.05 | 4.85 |
| C100 | 0.7 | 0.73 | 0.32 | 0.22 |
| C200 | 0.2 | 0.31 | 0.10 | 0.95 |

TABLE IV
SYNTHETIC DATA: THE SHAPE PARAMETER OF THE tanh FUNCTION DETERMINED BY THE RS METHOD

| Data | $\lambda_{real}$ | $\lambda_{opt}$ | | |
|---|---|---|---|---|
| | | Without noise | $N(0, 0.01)$ | $N(0, 0.1)$ |
| C20 | 1 | 1 | 1.04 | 1.45 |
| C40 | 1 | 0.95 | 1.42 | 1.49 |
| C100 | 0.8 | 0.80 | 1.50 | 0.55 |
| C200 | 0.4 | 0.54 | 0.51 | 0.17 |

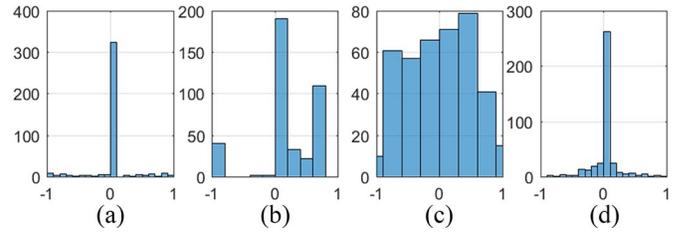

Fig. 2. Distribution of weight of the well-learned 20 nodes FCM with sigmoid function and Gaussian noise $N(0, 0.01)$. (a) Simulated. (b) NHL. (c) PSO. (d) LEFCM.

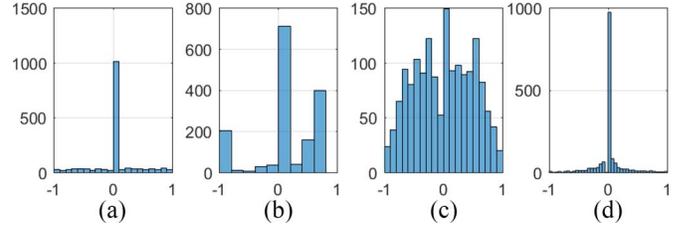

Fig. 3. Distribution of weight of the well-learned 40 nodes FCM with sigmoid function and Gaussian noise $N(0, 0.01)$. (a) Simulated. (b) NHL. (c) PSO. (d) LEFCM.

For the synthetic noise data experiments, the simulated FCMs come from Table II. The training response sequences are produced by these FCMs, and have the Gaussian noise $N(0, 0.01)$ and $N(0, 0.1)$ added, respectively. The hyperparameters of the LEFCM are optimized by the RS method. The experimental results of the quantitative comparisons are reported in Tables V–VIII. Analyzing the entries of these tables, the LEFCM receives top scores in 4 out of 5 evaluation metrics. It is clearly indicated that the performance of the LEFCM remarkably outperforms other learning methods.

In comparison with the other learning methods, the LEFCM generates the minimum *Data error* and *Out of sample error* for both small-scale and large-scale FCMs, even at orders of magnitude superior to PSO and NHL. It means that the response sequences that are generated by the FCMs that are learned by the proposed method are more approximate to the target sequences that are generated by the simulated FCMs. For example, in the synthetic noise data experiments with a sigmoid function, the average *Data error* is 0.0008 for the LEFCM, 0.0046 for the PSO, and 0.0068 for the NHL algorithm. The average *Out of sample error* is 0.0419 for the LEFCM, 0.1311 for the PSO, and 0.5041 for the NHL algorithm. The LEFCM reduces the mean *Data error* and *Out of sample error* rate by 83% and 68%, respectively, when compared with PSO, and 88% and 92%, respectively, when compared with the NHL. Meanwhile for the tanh function, the average *Data error* is 0.0051 for the LEFCM, 0.0245 for the PSO, and 0.0101 for the NHL algorithm. The average *Out of sample error* is 0.5423 for the LEFCM, 0.8403 for the PSO, and 0.9441 for the NHL algorithm. Likewise, the LEFCM reduces the mean *Data error* and *Out of sample error* rates by 79% and 35%, respectively, when compared with PSO, and 50% and 43%, respectively, when compared with the NHL. The lower *Data error* and *Out of sample error* indicate that the proposed method has better reconstruction and generalization abilities and a strong noise suppression ability.

Meanwhile, the LEFCM also provides the well-learned FCMs with the minimum *Model error* and the maximum SS Mean. For the experiments with the sigmoid function, the average *Model error* is 0.2025 for the LEFCM, 0.4647 for the PSO, and 0.5199 for the NHL algorithm. The average values of SS Mean is 0.58 for the LEFCM, 0.11 for the PSO, and 0.30 for the NHL algorithm. Compared with the PSO and NHL, the LEFCM reduces the mean *Model error* rate by 56% and 61%, respectively, and it increases the mean SS Mean rate by 427% and 93%, respectively. Whereas for the tanh function, the average *Model error* is 0.1216 for the LEFCM, 0.4344 for the PSO, and 0.6262 for the NHL algorithm, and the average SS Mean is 0.76 for the LEFCM, 0.13 for the PSO, and 0.15 for the NHL algorithm. Compared with PSO and NHL, the LEFCM reduces the mean *Model error* rate by 72% and 81%, respectively. In addition, it increases the mean SS Mean rate by 485% and 407%, respectively. The lower *Model error* and higher SS Mean show the better abilities of the LEFCM to estimate the weights and predict the existence of links, respectively.

With regard to the *Execution time*, the learning efficiency of the LEFCM is better than PSO, but it is worse than NHL. However, the NHL results in a high *Data error*, *Out of sample error*, and *Model error* and a low SS Mean in all cases. Although NHL is the fastest learning algorithm, it performs poorly in other metrics.

Further, the distributions of the weights that are generated by different learning methods are also reported in Figs. 2–9. In these figures, the abscissa is the weights and the ordinate is the number of observations of a certain weight. Observing these figures, we can see that the values of the weights that are learned by other methods are spread all over the interval $[-1, 1]$, whereas the values of the weights that are learned by the LEFCM are sparse and the distribution of the weights is similar to a normal distribution. A reasonable distribution of the weights can also reflect the good performance of the learning method.

TABLE V
EXPERIMENTAL RESULTS FOR SYNTHETIC DATA WITH GAUSSIAN NOISE $N(0, 0.01)$ AND SIGMOID FUNCTION

| Data | Algorithms | Hyper-parameters | | | Data error | Out of sample error | Model error | SS Mean | Time |
|---|---|---|---|---|---|---|---|---|---|
| | | $\alpha_{opt}$ | $\beta_{opt}$ | $\lambda_{opt}$ | | | | | |
| C20 ($\lambda_{real}$=5) | LEFCM | 0.0988 | 0.0395 | 5.03 | 0.0003±0.0000 | 0.0164±0.0022 | 0.0943±0.0047 | 0.74±0.02 | 5.8±0.2 |
| | PSO | / | / | 5* | 0.0067±0.0024 | 0.2171±0.0806 | 0.4535±0.0241 | 0.09±0.01 | 50±8.3 |
| | NHL | / | / | 5* | 0.0063±0.0003 | 0.4967±0.0547 | 0.4225±0.0270 | 0.43±0.05 | 0.07±0.01 |
| C40 ($\lambda_{real}$=5) | LEFCM | 0.0288 | 0.0829 | 5.05 | 0.0011±0.0003 | 0.0525±0.0135 | 0.1522±0.0019 | 0.65±0.01 | 13.1±0.2 |
| | PSO | / | / | 5* | 0.0073±0.0006 | 0.1478±0.0087 | 0.4802±0.0087 | 0.12±0.01 | 114±12 |
| | NHL | / | / | 5* | 0.0115±0.0006 | 0.5267±0.0467 | 0.4898±0.0072 | 0.42±0.04 | 0.04±0.01 |
| C100 ($\lambda_{real}$=0.7) | LEFCM | 0.1806 | 0.0484 | 0.32 | 0.0001±0.0000 | 0.0117±0.0003 | 0.2017±0.0007 | 0.55±0.00 | 41.4±0.1 |
| | PSO | / | / | 0.7* | 0.0033±0.0001 | 0.1105±0.0075 | 0.4741±0.0026 | 0.11±0.00 | 98.1±1.5 |
| | NHL | / | / | 0.7* | 0.0056±0.0001 | 0.4992±0.0145 | 0.5047±0.0161 | 0.35±0.03 | 0.11±0.02 |
| C200 ($\lambda_{real}$=0.2) | LEFCM | 0.2109 | 0.0198 | 0.10 | 0.0001±0.0000 | 0.0082±0.0001 | 0.2059±0.0008 | 0.53±0.00 | 146±7.4 |
| | PSO | / | / | 0.2* | 0.0018±0.0001 | 0.0595±0.0026 | 0.4736±0.0034 | 0.12±0.00 | 213±3.1 |
| | NHL | / | / | 0.2* | 0.0038±0.0000 | 0.4978±0.0001 | 0.6611±0.0005 | 0.00±0.00 | 0.27±0.03 |

[1] "*" indicates that the values of the hyper-parameters are predefined, other are obtained by Random search.
[2] "/" means that the parameter is not involved.

TABLE VI
EXPERIMENTAL RESULTS FOR SYNTHETIC DATA WITH GAUSSIAN NOISE $N(0, 0.1)$ AND SIGMOID FUNCTION

| Data | Algorithms | Hyper-parameters | | | Data error | Out of sample error | Model error | SS Mean | Time |
|---|---|---|---|---|---|---|---|---|---|
| | | $\alpha_{opt}$ | $\beta_{opt}$ | $\lambda_{opt}$ | | | | | |
| C20 ($\lambda_{real}$=5) | LEFCM | 0.0487 | 0.0643 | 5.40 | 0.0011±0.0001 | 0.0415±0.0055 | 0.1544±0.0025 | 0.59±0.01 | 8.9±0.1 |
| | PSO | / | / | 5* | 0.0073±0.0015 | 0.2390±0.0459 | 0.4238±0.0140 | 0.11±0.01 | 83.1±4.7 |
| | NHL | / | / | 5* | 0.0062±0.0005 | 0.4891±0.0668 | 0.4295±0.0288 | 0.41±0.07 | 0.07±0.02 |
| C40 ($\lambda_{real}$=5) | LEFCM | 0.1631 | 0.0708 | 4.85 | 0.0018±0.0001 | 0.0862±0.0226 | 0.2101±0.0029 | 0.55±0.01 | 14.1±0.3 |
| | PSO | / | / | 5* | 0.0057±0.0013 | 0.1140±0.0161 | 0.4867±0.0053 | 0.11±0.01 | 124±2.6 |
| | NHL | / | / | 5* | 0.0112±0.0006 | 0.5237±0.0497 | 0.4944±0.0107 | 0.42±0.04 | 0.05±0.01 |
| C100 ($\lambda_{real}$=0.7) | LEFCM | 0.1744 | 0.0759 | 0.22 | 0.0011±0.0000 | 0.0349±0.0014 | 0.3492±0.0054 | 0.50±0.01 | 42.1±0.6 |
| | PSO | / | / | 0.7* | 0.0031±0.0004 | 0.1015±0.0087 | 0.4706±0.0023 | 0.12±0.01 | 98.6±0.6 |
| | NHL | / | / | 0.7* | 0.0057±0.0001 | 0.5014±0.0163 | 0.5028±0.0164 | 0.35±0.03 | 0.11±0.03 |
| C200 ($\lambda_{real}$=0.2) | LEFCM | 0.1492 | 0.0625 | 0.95 | 0.0006±0.0000 | 0.0840±0.0030 | 0.2525±0.0019 | 0.50±0.00 | 134±1.4 |
| | PSO | / | / | 0.2* | 0.0018±0.0000 | 0.0597±0.0012 | 0.4766±0.0026 | 0.12±0.00 | 215±4.5 |
| | NHL | / | / | 0.2* | 0.0039±0.0000 | 0.4978±0.0001 | 0.6545±0.0006 | 0.00±0.00 | 0.26±0.02 |

[1] "*" indicates that the values of the hyper-parameters are predefined, other are obtained by Random search.
[2] "/" means that the parameter is not involved.

TABLE VII
EXPERIMENTAL RESULTS FOR SYNTHETIC DATA WITH GAUSSIAN NOISE $N(0, 0.01)$ AND tanh FUNCTION

| Data | Algorithms | Hyper-parameters | | | Data error | Out of sample error | Model error | SS Mean | Time |
|---|---|---|---|---|---|---|---|---|---|
| | | $\alpha_{opt}$ | $\beta_{opt}$ | $\lambda_{opt}$ | | | | | |
| C20 ($\lambda_{real}$=1) | LEFCM | 0.0733 | 0.1233 | 1.04 | 0.0017±0.0003 | 0.1123±0.0487 | 0.0134±0.0048 | 0.98±0.01 | 6.2±0.3 |
| | PSO | / | / | 1* | 0.0199±0.0010 | 0.7537±0.0526 | 0.4163±0.0228 | 0.11±0.01 | 25.4±9.3 |
| | NHL | / | / | 1* | 0.0084±0.0027 | 0.7776±0.2743 | 0.4494±0.0182 | 0.18±0.02 | 0.06±0.01 |
| C40 ($\lambda_{real}$=1) | LEFCM | 0.0387 | 0.1095 | 1.42 | 0.0069±0.0003 | 0.5981±0.0363 | 0.0629±0.0009 | 0.91±0.00 | 13.2±0.3 |
| | PSO | / | / | 1* | 0.0277±0.0021 | 0.8798±0.0122 | 0.4610±0.0104 | 0.13±0.02 | 44.5±0.8 |
| | NHL | / | / | 1* | 0.0099±0.0001 | 0.9834±0.0126 | 0.7784±0.0143 | 0.14±0.05 | 0.27±0.04 |
| C100 ($\lambda_{real}$=0.8) | LEFCM | 0.0477 | 0.2734 | 1.50 | 0.0079±0.0004 | 0.7072±0.0368 | 0.1129±0.0010 | 0.75±0.01 | 37.9±0.8 |
| | PSO | / | / | 0.8* | 0.0268±0.0003 | 0.9314±0.0133 | 0.4628±0.0022 | 0.12±0.01 | 45.0±0.3 |
| | NHL | / | / | 0.8* | 0.0132±0.0000 | 1.0062±0.0085 | 0.6915±0.0096 | 0.13±0.01 | 0.13±0.02 |
| C200 ($\lambda_{real}$=0.4) | LEFCM | 0.1724 | 0.1935 | 0.51 | 0.0005±0.0004 | 0.5712±0.0112 | 0.1400±0.0007 | 0.66±0.00 | 121.6±0.7 |
| | PSO | / | / | 0.4* | 0.0248±0.0004 | 0.8469±0.0085 | 0.4545±0.0184 | 0.12±0.02 | 114.1±4.2 |
| | NHL | / | / | 0.4* | 0.0089±0.0000 | 0.9952±0.0040 | 0.5836±0.0032 | 0.14±0.00 | 0.24±0.02 |

[1] "*" indicates that the values of the hyper-parameters are predefined, other are obtained by Random search.
[2] "/" means that the parameter is not involved.

TABLE VIII
EXPERIMENTAL RESULTS FOR SYNTHETIC DATA WITH GAUSSIAN NOISE $N(0, 0.1)$ AND tanh FUNCTION

| Data | Algorithms | Hyper-parameters | | | Data error | Out of sample error | Model error | SS Mean | Time |
|---|---|---|---|---|---|---|---|---|---|
| | | $\alpha_{opt}$ | $\beta_{opt}$ | $\lambda_{opt}$ | | | | | |
| C20 ($\lambda_{real}$=1) | LEFCM | 0.1516 | 0.3455 | 1.45 | 0.0033±0.0002 | 0.2448±0.0446 | 0.0873±0.0029 | 0.76±0.02 | 5.8±0.1 |
| | PSO | / | / | 1* | 0.0161±0.0010 | 0.6105±0.0497 | 0.3424±0.0133 | 0.19±0.02 | 33.0±0.3 |
| | NHL | / | / | 1* | 0.0088±0.0022 | 0.8054±0.2290 | 0.4579±0.0202 | 0.18±0.07 | 0.06±0.01 |
| C40 ($\lambda_{real}$=1) | LEFCM | 0.0845 | 0.2139 | 1.49 | 0.0081±0.0003 | 0.7240±0.0227 | 0.0905±0.0011 | 0.76±0.01 | 13.0±0.1 |
| | PSO | / | / | 1* | 0.0291±0.0029 | 0.9262±0.0371 | 0.4168±0.0097 | 0.13±0.01 | 44.7±1.1 |
| | NHL | / | / | 1* | 0.0099±0.0000 | 0.9834±0.0126 | 0.7837±0.0194 | 0.15±0.05 | 0.29±0.07 |
| C100 ($\lambda_{real}$=0.8) | LEFCM | 0.0275 | 0.3832 | 0.55 | 0.0085±0.0003 | 0.7558±0.0419 | 0.1958±0.0028 | 0.63±0.00 | 42.5±0.9 |
| | PSO | / | / | 0.8* | 0.0273±0.0002 | 0.9297±0.0089 | 0.4577±0.0017 | 0.11±0.00 | 44.5±0.4 |
| | NHL | / | / | 0.8* | 0.0131±0.0000 | 1.0062±0.0085 | 0.6822±0.0105 | 0.13±0.01 | 0.12±0.01 |
| C200 ($\lambda_{real}$=0.4) | LEFCM | 0.1539 | 0.4793 | 0.17 | 0.0039±0.0000 | 0.6249±0.0159 | 0.2698±0.0021 | 0.60±0.00 | 135.7±1.2 |
| | PSO | / | / | 0.4* | 0.0243±0.0004 | 0.8438±0.0046 | 0.4633±0.0039 | 0.12±0.01 | 110.7±0.7 |
| | NHL | / | / | 0.4* | 0.0089±0.0000 | 0.9952±0.0040 | 0.5830±0.0040 | 0.14±0.00 | 0.23±0.02 |

[1] "*" indicates that the values of the hyper-parameters are predefined, other are obtained by Random search.
[2] "/" means that the parameter is not involved.

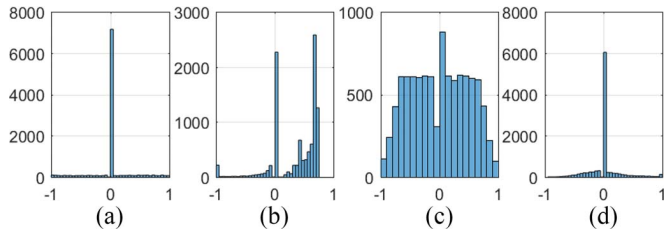

Fig. 4. Distribution of weight of the well-learned 100 nodes FCM with sigmoid function and Gaussian noise $N(0, 0.01)$. (a) Simulated. (b) NHL. (c) PSO. (d) LEFCM.

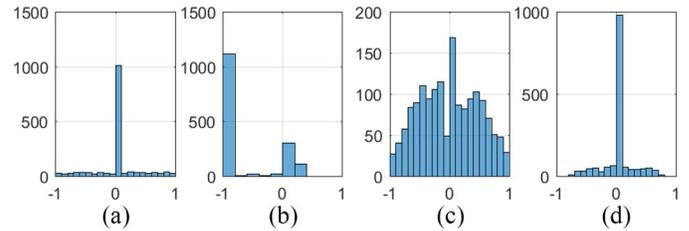

Fig. 7. Distribution of weight of the well-learned 40 nodes FCM with tanh function and Gaussian noise $N(0, 0.01)$. (a) Simulated. (b) NHL. (c) PSO. (d) LEFCM.

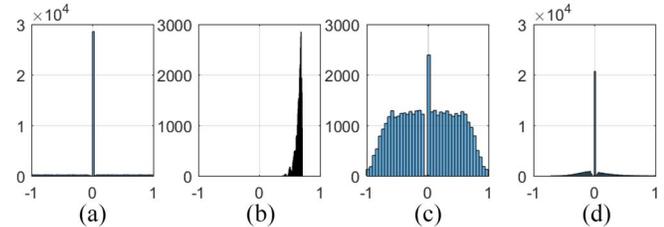

Fig. 5. Distribution of weight of the well-learned 200 nodes FCM with sigmoid function and Gaussian noise $N(0, 0.01)$. (a) Simulated. (b) NHL. (c) PSO. (d) LEFCM.

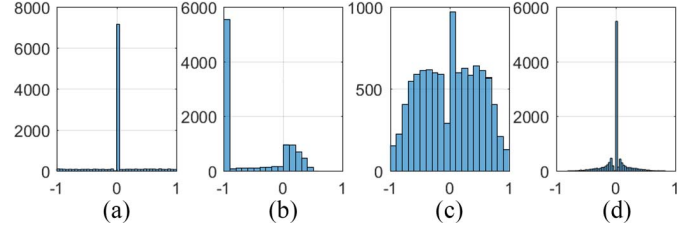

Fig. 8. Distribution of weight of the well-learned 100 nodes FCM with tanh function and Gaussian noise $N(0, 0.01)$. (a) Simulated. (b) NHL. (c) PSO. (d) LEFCM.

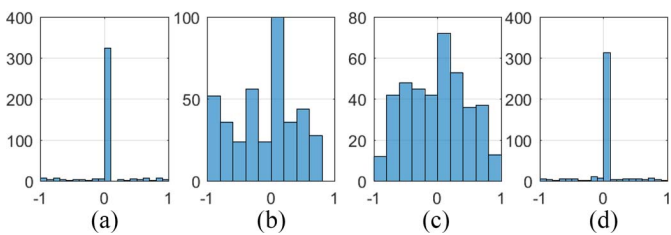

Fig. 6. Distribution of weight of the well-learned 20 nodes FCM with tanh function and Gaussian noise $N(0, 0.01)$. (a) Simulated. (b) NHL. (c) PSO. (d) LEFCM.

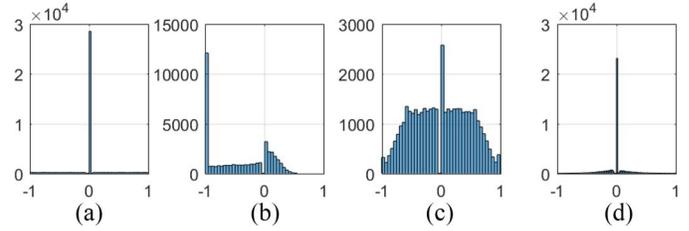

Fig. 9. Distribution of weight of the well-learned 200 nodes FCM with tanh function and Gaussian noise $N(0, 0.01)$. (a) Simulated. (b) NHL. (c) PSO. (d) LEFCM.

The above comparisons also reveal that LEFCM has a strongly robust learning ability in noisy environments. The reason is that the learning problem of the FCM is converted into a constrained convex optimization problem in which the least square item makes the produced weights have a better ability to tolerate noise, the $L_1$ regularization term makes the produced weights become sparse, and the maximum entropy term makes the distribution of the produced weights more reasonable.

### C. Real-World Data

Two real small-scale FCMs from [43] are used to further test the performance of the LEFCM. One is a model of factors

TABLE IX
EXPERIMENTAL RESULTS FOR REAL DATA ADDING GAUSSIAN NOISE $N(0, 0.01)$

| Data | Algorithms | Hyper-parameters | | | Data error | Out of sample error | Model error | SS Mean | Time |
|---|---|---|---|---|---|---|---|---|---|
| | | $\alpha_{opt}$ | $\beta_{opt}$ | $\lambda_{opt}$ | | | | | |
| C13 | LEFCM | 0.1014 | 0.0904 | 5.26 | 0.0007±0.0001 | 0.0156±0.0020 | 0.0412±0.0038 | 0.85±0.03 | 2.9±0.1 |
| | PSO | / | / | 5* | 0.0130±0.0027 | 0.0826±0.0263 | 0.3884±0.0113 | 0.11±0.07 | 11.5±1.0 |
| | NHL | / | / | 5* | 0.0179±0.0030 | 0.4692±0.1095 | 0.5444±0.0157 | 0.39±0.07 | 0.02±0.00 |
| C24 | LEFCM | 0.1967 | 0.1556 | 5.28 | 0.0003±0.0000 | 0.0080±0.0013 | 0.0388±0.0068 | 0.90±0.03 | 5.7±0.1 |
| | PSO | / | / | 5* | 0.0051±0.0019 | 0.0257±0.0012 | 0.4150±0.0144 | 0.13±0.02 | 18.5±0.2 |
| | NHL | / | / | 5* | 0.0152±0.0014 | 0.5258±0.0945 | 0.6031±0.0103 | 0.36±0.05 | 0.02±0.01 |

[1] "*" indicates that the values of the hyper-parameters are predefined, other are obtained by Random search.
[2] "/" means that the parameter is not involved.

TABLE X
EXPERIMENTAL RESULTS FOR REAL DATA ADDING GAUSSIAN NOISE $N(0, 0.1)$

| Data | Algorithms | Hyper-parameters | | | Data error | Out of sample error | Model error | SS Mean | Time |
|---|---|---|---|---|---|---|---|---|---|
| | | $\alpha_{opt}$ | $\beta_{opt}$ | $\lambda_{opt}$ | | | | | |
| C13 | LEFCM | 0.1348 | 0.1010 | 4.51 | 0.0007±0.0001 | 0.0141±0.0027 | 0.0434±0.0037 | 0.85±0.03 | 2.9±0.1 |
| | PSO | / | / | 5* | 0.0115±0.0022 | 0.0714±0.0091 | 0.3887±0.0297 | 0.13±0.03 | 11.3±0.2 |
| | NHL | / | / | 5* | 0.0177±0.0029 | 0.4692±0.1095 | 0.5454±0.0160 | 0.39±0.07 | 0.02±0.00 |
| C24 | LEFCM | 0.2535 | 0.1728 | 4.62 | 0.0019±0.0000 | 0.0177±0.0025 | 0.0792±0.0069 | 0.64±0.09 | 5.7±0.0 |
| | PSO | / | / | 5* | 0.0035±0.0011 | 0.0279±0.0044 | 0.4201±0.0151 | 0.12±0.01 | 14.5±5.5 |
| | NHL | / | / | 5* | 0.0146±0.0012 | 0.5095±0.0778 | 0.6070±0.0103 | 0.36±0.04 | 0.02±0.01 |

[1] "*" indicates that the values of the hyper-parameters are predefined, other are obtained by Random search.
[2] "/" means that the parameter is not involved.

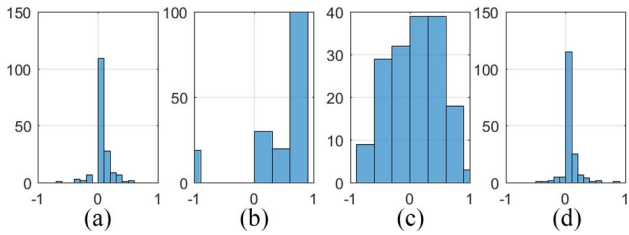

Fig. 10. Distribution of weight of the well-learned 13 nodes FCM with Gaussian noise $N(0, 0.01)$. (a) Real FCMs. (b) NHL. (c) PSO. (d) LEFCM.

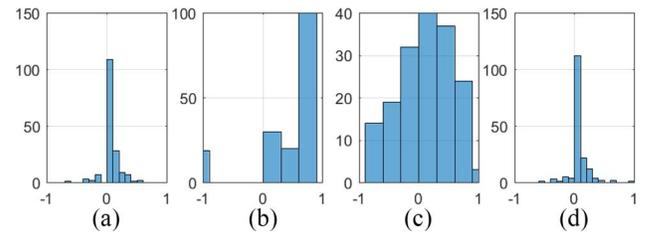

Fig. 12. Distribution of weight of the well-learned 13 nodes FCM with Gaussian noise $N(0, 0.1)$. (a) Real FCMs. (b) NHL. (c) PSO. (d) LEFCM.

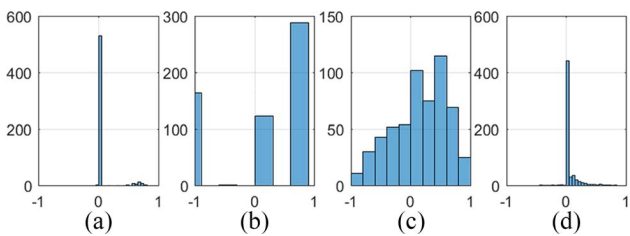

Fig. 11. Distribution of weight of the well-learned 24 nodes FCM with Gaussian noise $N(0, 0.01)$. (a) Real FCMs. (b) NHL. (c) PSO. (d) LEFCM.

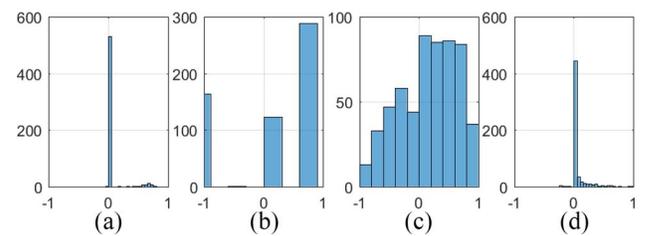

Fig. 13. Distribution of weight of the well-learned 24 nodes FCM with Gaussian noise $N(0, 0.1)$. (a) Real FCMs. (b) NHL. (c) PSO. (d) LEFCM.

affecting the slurry rheology with 13 nodes [44], and the other is a model of factors in the adoption of educational software in schools with 24 nodes [45]. These two models are simulated from random initial vectors and have the Gaussian noise added, and the experimental procedures are implemented in the same way as in the previous section. Since the values of the nodes of these networks are positive, the activation functions that are used in these FCMs are set as all sigmoid functions with the value of $\lambda$ set to 5. Furthermore, ten independent response sequences with noise are generated for every trial where each sequence contains ten steps. The quantitative comparison results are reported in Tables IX and X. In comparison with the PSO and NHL algorithms, the LEFCM decreases the overall *Data error* and *Out of sample error* rates by at least 45% and 36%, respectively. Likewise, the LEFCM decreases the overall *Model error* rate by at least 80% and increases the overall SS Mean rate by at least 77%. As expected, the LEFCM is much better than other leaning methods.

Figs. 10–13 also show the weight distributions of the well-learned FCMs for different learning methods. In these figures,

TABLE XI
EXPERIMENTAL RESULTS FOR DREAM4 WITH 100 GENES

| Data | Algorithms | Hyper-parameters | | | Data error | Out of sample error | SS Mean | Time |
|---|---|---|---|---|---|---|---|---|
| | | $\alpha_{opt}$ | $\beta_{opt}$ | $\lambda_{opt}$ | | | | |
| Dream4-100-1 | LEFCM | 0.2072 | 0.4646 | 0.21 | 0.0010±0.0000 | 0.0927±0.0078 | 0.70±0.01 | 59.2±0.6 |
| | PSO | / | / | 0.7* | 0.0012±0.0000 | 0.1262±0.0091 | 0.11±0.00 | 144±1.3 |
| | NHL | / | / | 0.7* | 0.0054±0.0000 | 0.6822±0.0098 | 0.06±0.01 | 0.16±0.01 |
| Dream4-100-2 | LEFCM | 0.2304 | 0.2312 | 0.24 | 0.0010±0.0000 | 0.0976±0.0100 | 0.58±0.01 | 59.5±0.9 |
| | PSO | / | / | 0.7* | 0.0022±0.0000 | 0.2355±0.0100 | 0.12±0.00 | 144±0.7 |
| | NHL | / | / | 0.7* | 0.0049±0.0000 | 0.6175±0.0079 | 0.01±0.01 | 0.17±0.01 |
| Dream4-100-3 | LEFCM | 0.0662 | 0.3093 | 0.31 | 0.0010±0.0000 | 0.0872±0.0088 | 0.62±0.01 | 58.2±0.6 |
| | PSO | / | / | 0.7* | 0.0023±0.0001 | 0.2578±0.0100 | 0.12±0.00 | 144±0.2 |
| | NHL | / | / | 0.7* | 0.0051±0.0000 | 0.6417±0.0095 | 0.03±0.01 | 0.16±0.01 |
| Dream4-100-4 | LEFCM | 0.2292 | 0.1471 | 0.83 | 0.0011±0.0001 | 0.1095±0.0136 | 0.62±0.01 | 58.8±0.5 |
| | PSO | / | / | 0.7* | 0.0020±0.0000 | 0.2180±0.0063 | 0.11±0.00 | 144±0.8 |
| | NHL | / | / | 0.7* | 0.0051±0.0000 | 0.6513±0.0102 | 0.01±0.01 | 0.17±0.01 |
| Dream4-100-5 | LEFCM | 0.1756 | 0.2500 | 0.64 | 0.0010±0.0000 | 0.0894±0.0122 | 0.60±0.01 | 59.0±0.5 |
| | PSO | / | / | 0.7* | 0.0021±0.0000 | 0.2333±0.0039 | 0.11±0.01 | 144±0.6 |
| | NHL | / | / | 0.7* | 0.0052±0.0000 | 0.6612±0.0077 | 0.01±0.01 | 0.16±0.01 |

[1] "*" indicates that the values of the hyper-parameters are predefined, other are obtained by Random search.
[2] "/" means that the parameter is not involved.

the abscissa is the weight values and the ordinate is the number of observations of a certain weight. They clearly show that the weight matrices that are learned by the LEFCM are most similar to the real ones.

To verify the ability of the LEFCM to learn real large-scale networks, the DREAM4 data from the *in silico* network challenge [32], [46] is used to reconstruct the GRNs with 100 genes. The DREAM4 data with 100 genes contains 5 different networks. Each network provides 10 time series that are observed under different perturbations and each time series has 21 time points with noise. Here, our objective is to construct the FCMs with 100 nodes (the activation function of these FCMs is selected as the sigmoid function since the values of the nodes of the GRNs are all positive) to express these GRNs, and explore whether there are links between the genes in light of the corresponding groups of the time-series data. In the experiments, nine time series are randomly selected as the training set and the remaining one is used as the testing set for each network. When the FCMs are learned on the basis of the training set by the LEFCM, PSO, and NHL, respectively, the corresponding *Data error* is calculated for the training set while the *Out of sample error* is calculated for the testing set. Since the networks in the DREAM4 data are binary networks, the *Model error* is unsuitable for the comparison. The corresponding experimental results are reported in Table XI. The comparison results also show that the proposed method has a better learning ability than the existing methods.

## V. CONCLUSION

Learning FCMs from data is always a difficult task, especially, when encountered lots of noisy data. To deal with the learning problem of FCMs with noisy data, a simple and efficient method is proposed in this article. The crux of the proposed method is that the learning problem of the FCMs is considered as a convex optimization problem with constraints. Experiments completed with synthetic noisy data and publicly available data with noise demonstrate that the proposed method can not only produce the weight with better distribution rapidly and efficiently when learning small-scale FCMs as well as large-scale FCMs but also exhibit the ability to become robust against noise.

This article can be regarded as a starting point for supporting further studies. There are several main and promising directions that worth pursuing.

1) *Exploring the Advanced Hybrid Learning Algorithms:* As presented in Section III, some prior knowledge that are implied in data and reflect information of the structure of the FCM are not exploited in the proposed method. Based on this, a hybrid learning algorithm of FCM is developed in the way in which the knowledge associated with the structure of the FCM are mined from historical data by means of some existing methods, such as mutual information, etc., and introduced into the proposed method, which could further reduce the learning time of the FCMs and enhance the ability to develop high quality FCMs.

2) *Developing Distributed Model of FCMs for the Description of the Complex System:* A complex system, in general, consists of many subsystems involved a large number of variables. It is difficult to be accurately described by a single FCM. Here, the idea of granular computing can be borrowed. We first directly invoke the proposed method to construct a series of local FCMs to describe the individual subsystems. Subsequently, those local FCMs are connected in terms of the relationship between the corresponding subsystems, and the corresponding connection weights can be learned by again invoking the proposed method. Thus, a global FCM that is used to describe the entire complex system is formed.

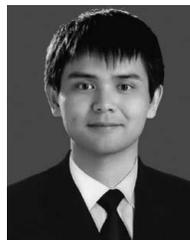

**Guoliang Feng** received the M.S. degree in control theory and control engineering from Northeast Electric Power University, Jilin, China, in 2011. He is currently pursuing the Ph.D. degree with the School of Control Science and Engineering, Dalian University of Technology, Dalian, China.

He joined the School of Automation Engineering, Northeast Electric Power University in 2011. His current research interests include fuzzy cognitive maps and granular computing.

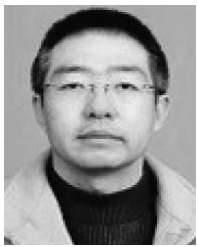

**Wei Lu** received the M.S. degree and the Ph.D. degree in control theory and control engineering from the Dalian University of Technology, Dalian, China, in 2004 and 2015, respectively.

In 2004, he joined the Dalian University of Technology, where he is currently an Associate Professor with the School of Control Science and Engineering. His current research interests include computational intelligence, fuzzy modeling and granular computing, knowledge discovery and data mining, and fuzzy intelligent systems.

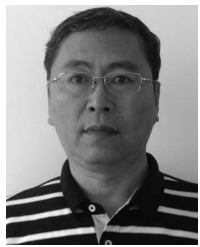

**Jianhua Yang** received the M.S. degree from Jilin University, Changchun, China, in 1987.

He joined the Dalian University of Technology, Dalian, China, in 1988, where he is currently a Professor with the School of Control Science and Engineering. His current research interests include intelligent control technologies, data acquisition, and low-cost integrated automation systems.

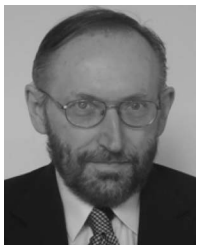

**Witold Pedrycz** (F'98) received the M.Sc., Ph.D., and D.Sc. degrees from the Silesian University of Technology, Gliwice, Poland.

He is a Professor and the Canada Research Chair of computational intelligence with the Department of Electrical and Computer Engineering, University of Alberta, Edmonton, AB, Canada. He is also with the Systems Research Institute, Polish Academy of Sciences, Warsaw, Poland, where he was elected as a Foreign Member in 2009, and with the Department of Electrical and Computer Engineering, Faculty of Engineering, King Abdulaziz University, Jeddah, Saudi Arabia. He has authored 17 research monographs and edited volumes covering various aspects of computational intelligence, data mining, and software engineering. His current research interests include computational intelligence, fuzzy modeling and granular computing, knowledge discovery and data science, fuzzy control, pattern recognition, knowledge-based neural networks, relational computing, and software engineering. He has published numerous papers in the above areas.

Prof. Pedrycz was a recipient of the Prestigious Norbert Wiener Award from the IEEE Systems, Man, and Cybernetics Society in 2007, the IEEE Canada Computer Engineering Medal, the Cajastur Prize for Soft Computing from the European Centre for Soft Computing, the Killam Prize, and the Fuzzy Pioneer Award from the IEEE Computational Intelligence Society. He is currently the Editor-in-Chief of *Information Sciences* and *WIREs Data Mining and Knowledge Discovery* (Wiley), and the Co-Editor-in-Chief of the *International Journal of Granular Computing* (Springer) and the *Journal of Data Information and Management* (Springer). He also serves on an Advisory Board of the IEEE TRANSACTIONS ON FUZZY SYSTEMS. He is vigorously involved in editorial activities. In 2012, he was elected as a fellow of the Royal Society of Canada.

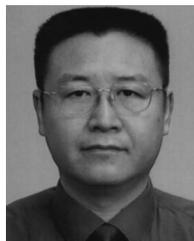

**Xiaodong Liu** received the B.S. degree in mathematics from Northeastern Normal University, Changchun, China, in 1986, the M.S. degree in mathematics from Jilin University, Jilin, China, in 1989, and the Ph.D. degree in control theory and control engineering from Northeastern University, Shenyang, China, in 2003.

He is currently a Professor with the School of Control Science and Engineering, Dalian University of Technology, Dalian, China, and the Department of Applied Mathematics, Dalian Maritime University, Dalian. He has proposed the axiomatic fuzzy set theory and has coauthored four books. His current research interests include knowledge discovery and representation, axiomatic fuzzy sets, fuzzy control, fuzzy sets and systems, artificial intelligence, machine learning, pattern recognition and hitch diagnosis, time series analysis, stochastic processes, and financial mathematics.